\documentclass[runningheads]{llncs}
\usepackage[T1]{fontenc}
\usepackage{graphicx,lipsum,afterpage,subcaption}
\usepackage{algorithm}
\usepackage{algorithmic}
\usepackage{times}
\usepackage{latexsym}
\usepackage{makecell}

\usepackage{amsmath,amsthm}
\usepackage{amssymb}
\usepackage{soul,xcolor}
\usepackage{siunitx}
\usepackage{url}
\usepackage{booktabs}
\usepackage{hyperref}
\usepackage{multirow}
\usepackage{lipsum}
\usepackage{newfloat}
\usepackage{listings}
\usepackage{tablefootnote}

\begin{document}
\title{{\em Yes, this is what I was looking for!} Towards Multi-modal Medical Consultation Concern Summary Generation}

\titlerunning{Towards Multi-modal Medical Consultation Concern Summary Generation}
%
\author{Abhisek Tiwari\inst{1} \and
Shreyangshu Bera\inst{1}  \and
Sriparna Saha\inst{1} \and Pushpak Bhattacharyya\inst{2} \and Samrat Ghosh\inst{1}}
\authorrunning{Tiwari et al.}
\institute{Dept. of Computer Science and Engineering, Indian Institute of Technology Patna, India \and Dept. of Computer Science and Engineering, Indian Institute of Technology Bombay, India
\email{\{abhisek\_1921cs16, sriparna\}@iitp.ac.in, iamshreyangshu07@gmail.com, pb@cse.iitb.ac.in, samratghosh080@gmail.com}
}

\maketitle              
\begin{abstract}
Over the past few years, the use of the Internet for healthcare-related tasks has grown by leaps and bounds, posing a challenge in effectively managing and processing information to ensure its efficient utilization. During moments of emotional turmoil and psychological challenges, we frequently turn to the internet as our initial source of support, choosing this over discussing our feelings with others due to the associated social stigma. In this paper, we propose a new task of multi-modal medical concern summary ({\em MMCS}) generation, which provides a short and precise summary of patients' major concerns brought up during the consultation. Nonverbal cues, such as patients' gestures and facial expressions, aid in accurately identifying patients' concerns. Doctors also consider patients' personal information, such as age and gender, in order to describe the medical condition appropriately. Motivated by the potential efficacy of patients' personal context and visual gestures, we propose a transformer-based multi-task, multi-modal intent-recognition, and medical concern summary generation ({\em IR-MMCSG}) system. Furthermore, we propose a multitasking framework for intent recognition and medical concern summary generation for doctor-patient consultations. We construct the first multi-modal medical concern summary generation ({\em MM-MediConSummation}) corpus, which includes patient-doctor consultations annotated with medical concern summaries, intents, patient personal information, doctor's recommendations, and keywords. Our experiments and analysis demonstrate (a) the significant role of patients' expressions/gestures and their personal information in intent identification and medical concern summary generation, and (b) the strong correlation between intent recognition and patients' medical concern summary generation\footnote{The dataset and source code are available at \url{https://github.com/NLP-RL/MMCSG}}.

\keywords{Clinical Conversation \and Concern Summary \and Multi-modality \and Modality Fusion \and Multi-tasking \and Summary Generation.}
\end{abstract}

\section{Introduction} 
In the past few years, tele-health has grown immensely with the advancement of information \& communication technologies (ICTs) and artificial intelligence-based applications for healthcare activities \cite{nittari2020telemedicine}. With the COVID-19 pandemic, internet utilization for healthcare activities has reached its peak and has become a new normal \cite{wosik2020telehealth}. The outbreak has caused a striking 25\% increase\footnote{{{https://www.who.int/news/item/02-03-2022-covid-19-pandemic-triggers-25-increase-in-prevalence-of-anxiety-and-depression-worldwide}}} in anxiety and depression, which are severely straining the mental healthcare systems. Tele-health usage is being actively encouraged by healthcare providers, and patients are adopting it at the same pace. Consequently, a massive amount of medical data became available for the first time over the internet \cite{barnes2019conversation}. Thus, arranging this data efficiently is essential for proper referencing and facilitating their potential for reuse.
\vspace{-1.5em}
\begin{figure*}
    \centering
    \includegraphics[scale=0.45]{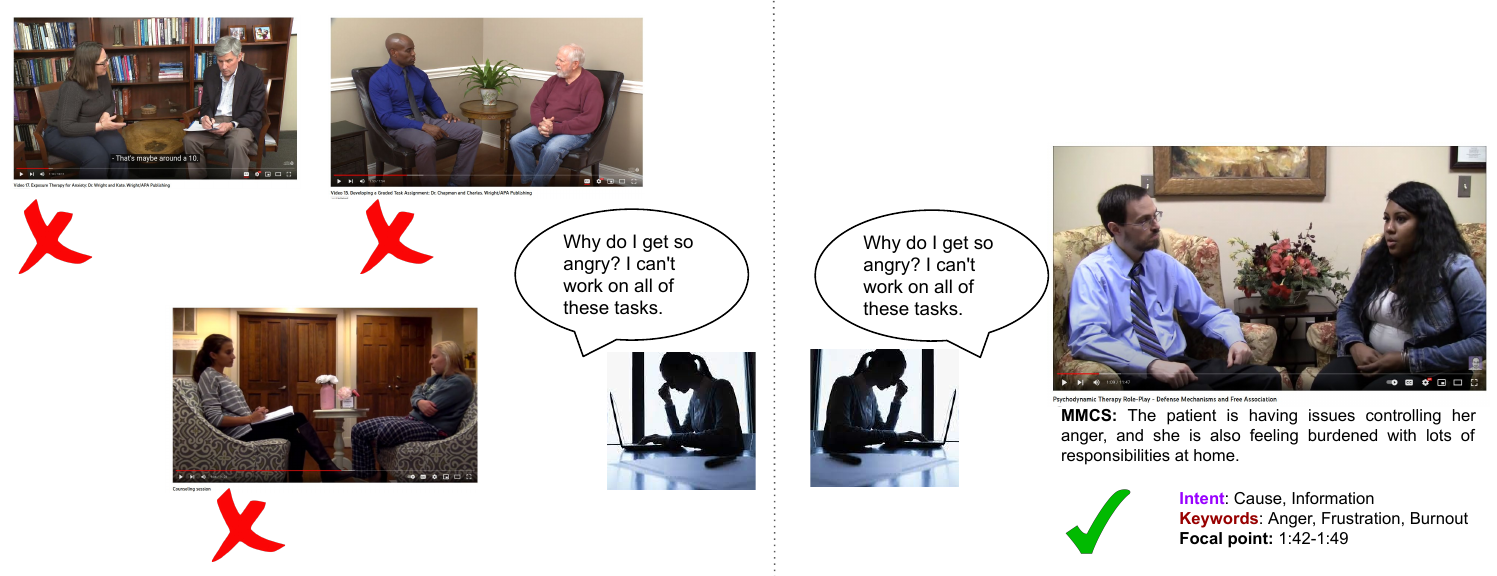}
    \vspace{-1em}
    \caption{Utility of multi-modal medical concern summary generation; generated medical concern summary for a video helps in selecting a proper video relevant to user (right side)}
    \vspace{-1.5em}
    \label{fig1}
\end{figure*}

  In moments of mental distress, individuals often turn to the internet to seek similar cases and recovery insights. However, they have to go through several lengthy videos before getting a relevant case. While many videos do provide summaries, these summaries combine information from both the patient and the doctor, rendering them ineffective as valuable references for finding pertinent cases that align closely with patients' main concerns (Fig. \ref{fig1}). Motivated by this, we propose a new task of generating Multi-modal Medical Concern Summary ({\em MMCS}). {\em MMCS} is a synopsis of a patient's key concerns discussed in a patient-doctor interaction. Its potential benefits extend to both patients and clinicians, serving various purposes, including (a) aiding in the organization of consultations and enhancing search ranking for reference by other patients, (b) facilitating follow-up recommendations, and (c) contributing to resource allocation and planning.

When we consult with doctors, they also consider our facial expressions and gestures to analyze our concerns and plan treatment accordingly. Visual expression and audio tone are also affected by user personality, so one behavior may be triggering for one personality while being normal for another. If the patient's key concern is known, it is easy to identify the patient's intent and vice-versa. Thus, we hypothesize there is a significant correlation between intent and medical concern summary. Moreover, we anticipate that visual and audio features are strongly associated with demographic information such as age and gender (context), and they can significantly influence the understanding of users' behavior and concerns with such context-attended features. Hence, we propose contextualized M-modality fusion, a new modality fusion technique that incorporates an adapter-based module into traditional transformer architecture to effectively infuse different modalities and end-user demographic information.

\hspace{-0.55cm}\textbf{Research Questions} The paper aims to investigate the following research questions: (a) Does the visual appearance and expression of a patient aid in determining his/her key medical concerns? (b) Is there a correlation between medical concern summary generation and user intent identification? (c) Can patients' personal characteristics, such as age and gender information, contribute to adequately understanding their medical issues and providing appropriate medical advice? \\
\hspace{-0.55cm}\textbf{Key Contributions} The key contributions of the work are four-fold, which are as follows:

\begin{itemize}
    \item We propose a new task of multi-modal medical concern summary ({\em MMCS}) generation, which generates a precise summary of key medical concerns  discussed during doctor-patient consultations, resulting in better content searchability and organization.
    \item  We first curated a Medical Concern Summary annotated multi-modal medical dataset named ({\em MM-MediConSummation}), which consists of patient-doctor counseling sessions annotated with a precise medical concern summary (MCS), intent, patient's personal attributes, doctor's key points, and keywords.
    \item We present a multitask, multi-modal intent recognition, and multi-modal medical concern summary generation ({\em IR-MMCSG}) model incorporated with an adapter-based contextualized M-modality fusion mechanism that evaluates audio tone and visual expression in conjunction with user demographic context. 
    \item  The proposed contextualized M-modality fusion incorporated {\em IR-MMCSG} outperforms existing state-of-the-art multi-modal text generation model across all evaluation metrics, including human evaluation. 
\end{itemize}

\section{Related Works} The work is mainly relevant to the following two research areas: Medical dialogue understanding and Medical dialogue summarization. We have summarized the relevant works in the following paragraphs.\\
\hspace{-0.55cm}\textbf{Medical Dialogue Understanding:}Diagnosis and treatment of diseases begin with patient-doctor interaction. Therefore, understanding patients' concerns from their utterances is critical to diagnosis and treatment outcomes \cite{enarvi2020generating}. The existing works on medical dialogue understanding can broadly be grouped into two categories: (a) pre-trained transformer-based joint intent and entity detection model \cite{weld2021survey} and (b) multi-label entity classification \cite{shi2020understanding}. The existing works on medical entity extraction \cite{dreisbach2019systematic,bay2021term,10.1145/3511808.3557296} are limited to medical attributes like symptoms and medicine, which are functions of an utterance rather than a conversation. The work \cite{jeblee2019extracting} proposed a pipelined machine learning system that identifies intent, symptoms, and disease from a patient-doctor conversation. In \cite{stratigi2020multidimensional}, the authors demonstrated that considering patients' education level, health literacy, and emotional state greatly improves the likelihood of recommending a relevant medical document. 

\hspace{-0.55cm}\textbf{Medical Dialogue Summarization:} Joshi et al., \cite{joshi2020dr} proposed a summarization model based on a pointer network generator. The model takes dialogues as input and generates a summary for each turn (doctor-patient) of the interaction. The work \cite{song2020summarizing} proposed a hierarchical encoder-tagger for summarizing medical patient-doctor conversations by identifying important utterances. Multi-modal summarization aims to generate coherent and important information from data having multiple modalities \cite{zhu2018msmo}. In the last few years, the main focus of multi-modal summarization has been to find co-relation among different modalities: text, audio, and image for video data \cite{apostolidis2021video}. In \cite{10.1145/3583780.3614870}, the authors investigated the impact of external medical knowledge infusion in text-based dialogue summarization and showed the efficacy of modeling the knowledge in generating medical terms preserving dialogue summaries. An important segment of a video is a subjective concern and may also vary among consumers. In \cite{huang2021gpt2mvs}, the authors have proposed a new task of user constraint-based summarization and proposed an attention mechanism to summarize the query-relevant content. Shang et al., \cite{shang2021multimodal} proposed a time-aware multi-modal transformer (TAMT) that leverages time stamps across image, text, and audio to generate an adequate and coherent video summary.

\section{Dataset} We first extensively scrutinized the existing benchmark video summarization datasets, and the summary is presented in Table \ref{ED}. The most relevant dataset for our proposed task is HOPE \cite{malhotra2022speaker}, which contains patient-doctor counseling sessions. The therapy sessions have been collected from open-source platforms such as YouTube, which are credible and authentic, recorded by psychiatrists and clinicians. The consultations cover therapy sessions for various mental distress like anxiety, depression, and post-traumatic stress disorder (PTSD). It contains only doctor-patient conversation transcripts and dialogue acts corresponding to each utterance. With the guidance of two psychiatrists and a doctor, we incorporate the following attributes into the existing HOPE dataset: medical concern summary, primary intent, secondary intent, doctor suggestion, focal point, and patient's personal context (gender and age group). 
\vspace{-1.5em}
 \begin{table*}[hbt!]
    \centering
    \caption{Characteristics of some of the most relevant existing medical dialogue datasets and comparison with the curated dataset}
    \scalebox{0.58}{
    \begin{tabular}{|l|p{5.5cm}|c|c|c|c|c|c|c|c|}
    \hline
      \textbf{Dataset} & \textbf{Description} & \textbf{Size} &  \textbf{Video}   & \textbf{Transcript} &  \textbf{Intent} & \textbf{MMCS}  & \textbf{Focal point}    &  \textbf{Keywords} & \textbf{Patient Personal Context} \\ \hline
     DAIC-WOZ \cite{gratch2014distress} & Patient- Psychiatrist conversation & 189 & $\checkmark$ &  $\checkmark$ & $\checkmark$ & $\times$  & $\times$  & $\times$ & $\times$ \\ 
      Dr. Summarize \cite{joshi2020dr} & Patient-Doctor conversations  & 1690 & $\times$ & $\checkmark$ & $\times$ & $\times$ & $\checkmark$ &  $\checkmark$ & $\times$ \\
      GPT3-ENS SS \cite{chintagunta2021medically} & Patient-Doctor conversations & 210 & $\times$  & $\checkmark$ &  $\times$ & $\checkmark$  &   $\times$ & $\checkmark$ &$\times$ \\
    CoDEC \cite{singh2023decode} &  Patient- Psychiatrist conversation &  30 &  $\checkmark$ & $\checkmark$ & $\times$ & $\times$ & $\times$ & $\times$ & $\times$ \\
    HOPE \cite{malhotra2022speaker} &  Patient- Psychiatrist conversation & 212 & $\checkmark$& $\checkmark$ & $\times$ & $\times$ & $\times$ & $\times$ & $\times$ \\
      {\em MM-MediConSummation}  & Patient- Psychiatrist conversation  & 467 & $\checkmark$  & $\checkmark$ & $\checkmark$ & $\checkmark$ & $\checkmark$ & $\checkmark$  & $\checkmark$ \\
     \hline
    \end{tabular}}
    \vspace{-3em}
    \label{ED}
\end{table*}
 
\subsection{Data Collection and Annotation}
We, along with the three medical experts, first analyzed a few therapy sessions of the HOPE dataset. The clinicians viewed a small subset of the dataset, curated a sample dataset of 25 therapy sessions, and annotated it with a multi-modal medical concern summary, patients' intent, and other crucial information, namely the summary of the doctor's suggestion, focal point, and keywords. Due to the dataset's limited number of samples, we expanded our search to include additional similar samples from open sources and platforms such as YouTube, focusing on credible channels. We employed two biology graduates and one medical student to scale up the dataset and annotation. We provided the sample dataset with a set of guidelines to annotate these tags. They first watched full videos and annotated the details based on a comprehensive understanding of all three modalities involved in a video, i.e., text, audio, and visual, for identification. The process of annotating a video involves crafting a medical concern summary (MCS), identifying the intent, and delineating a specific segment within the video that effectively conveys essential details regarding the patient's concerns and the doctor's recommendations. They manually generated transcripts for specific therapy sessions in cases where video captions were either unavailable or subject to restrictions. In order to ensure annotation agreement among the annotators, we calculated Fleiss' kappa \cite{fleiss1971measuring}. It was found to be 0.71, indicating a significant uniform annotation. The {\em MM-MediConSummation} statistics and its word cloud are provided in Table \ref{DStat} and Fig. \ref{WC}, respectively. 
\vspace{-2.5em}
\begin{table}%
\parbox{0.5\textwidth}{
\caption{{\em MM-MediConSummation} dataset statistics}
\scalebox{0.7}{
\begin{tabular}{|l|p{4.5cm}|}
    \hline
       \textbf{Entries} & \textbf{Value} \\
       \hline 
       \# of doctor-patient sessions   & 467 \\
        avg duration of video & 10 min 58 sec \\
       \# of utterances & 74473 \\
        avg. conversation length (utterance) & 159 \\
       \# of unique words & 13639\\ 
       \# of intents & 7 \\ 
       avg length of MMCS (\# of words) & 21 \\
       \# of age-groups & 6 \\ 
       tags & MMCS, keywords, doctor suggestion, focus points, patient gender, patient age, primary intent, secondary intent, transcript \\
       \hline
    \end{tabular}}
\label{DStat}}
\qquad
\begin{minipage}[c]{0.47\textwidth}%
\centering
    \includegraphics[scale=0.19]{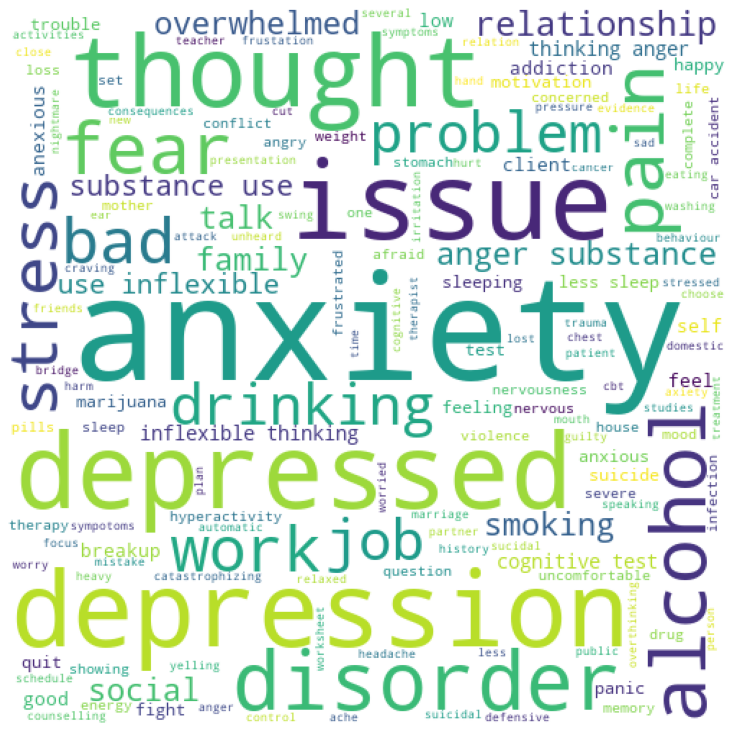}
\captionof{figure}{Word cloud of the {\em MM-MediConSummation} corpus}
\label{WC}
\end{minipage}
\vspace{-5.15em}
\end{table}
\subsection{Qualitative Aspects} 
To generate an adequate summary of patient concerns, we analyze different qualitative characteristics of therapy sessions and incorporate them accordingly. The different characteristics are analyzed and illustrated below.

\hspace{-0.55cm}\textbf{Role of Medical Concern Summary} Medical concern summary aims to aid online healthcare users in recognizing whether the content (particularly therapy session) contains the information they are looking for or not. For example, Fig. \ref{fig1} illustrates that the patient easily finds out a relevant video when the therapy sessions are tagged with {\em MMCS}, intent, and keywords. Additionally, we have labeled each session with a focal point, indicating the specific segment where the user articulates the primary purpose of the discussion.

\begin{figure}[hbt!]
\centering
\begin{minipage}{.46\columnwidth}
  \centering
  \includegraphics[scale=0.26]{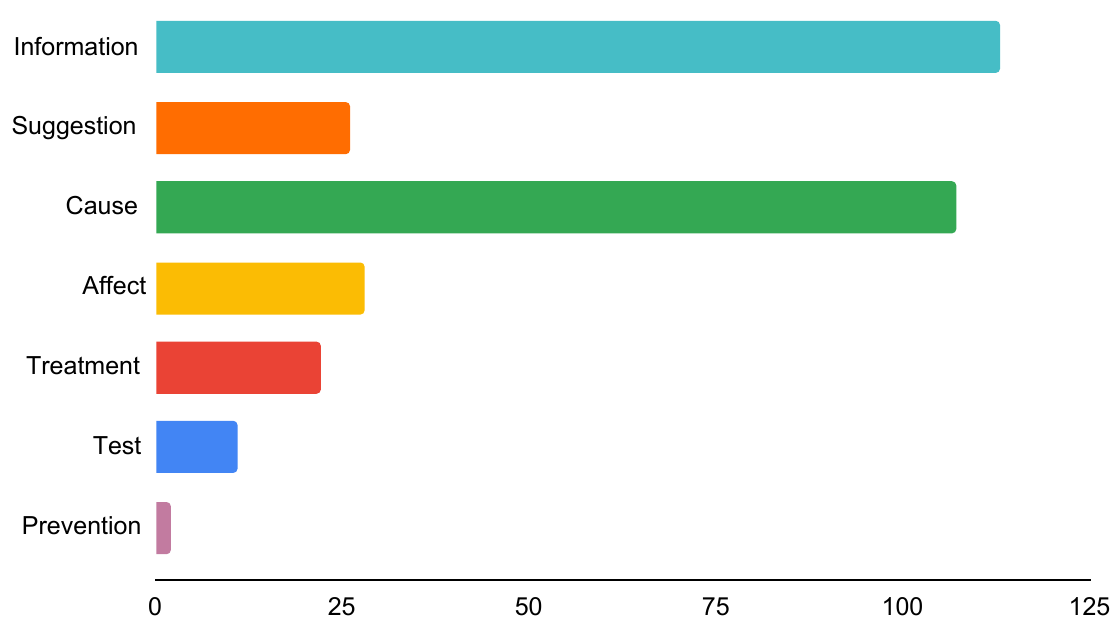}
  \vspace{-0.5em}
  \captionof{figure}{Primary intent distribution}
    \vspace{-1em}
  \label{INT1}
\end{minipage}%
\hspace{1.5em}
\begin{minipage}{.47\columnwidth}
  \centering
  \includegraphics[scale=0.24]{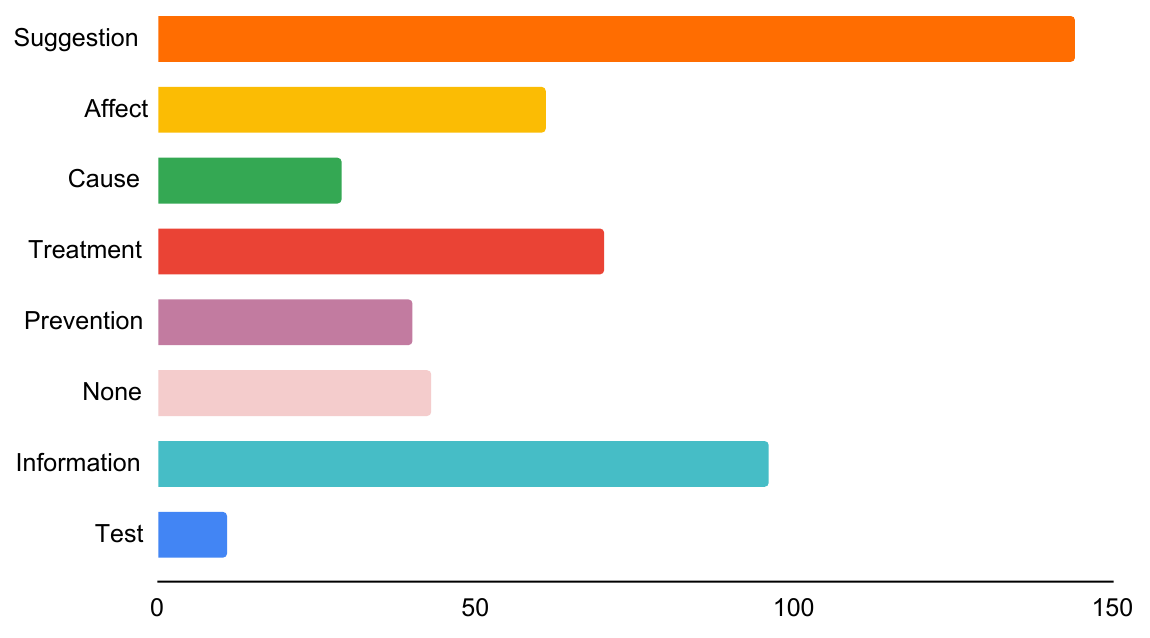}
   \vspace{-0.5em}
  \captionof{figure}{Secondary intent distribution}
  \vspace{-1em}
  \label{INT2}
\end{minipage}
\end{figure}

\hspace{-0.55cm}\textbf{Role of Intent} The user's intent in a video pertains to the purpose behind the patient's interaction with the clinician. It is desirable to comprehend users' intent to serve them effectively. For example, a user's goal might be to get a suggestion for a medical condition. We can use it to effectively locate the relevant span in the concerned video. Sometimes, we observed patients having multiple intentions for consultation; thus, we tagged primary and secondary intentions (Fig. \ref{INT1} and Fig. \ref{INT2}) to each session. We have used only primary intent for multi-tasking intent identification and MMCS generation.
\begin{figure}[hbt!]
\centering
\begin{minipage}{.46\columnwidth}
  \centering
  \includegraphics[scale=0.31]{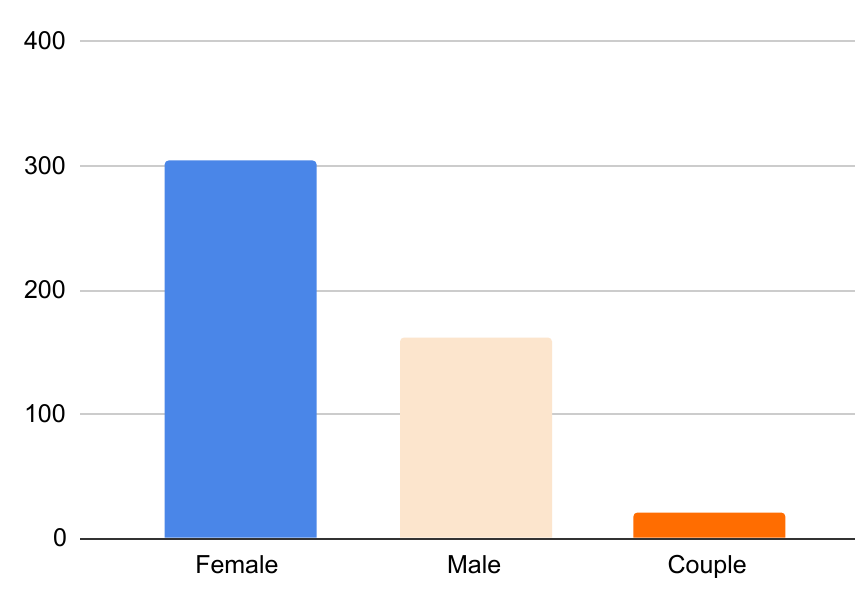}
  \captionof{figure}{Gender distribution}
  \label{GEN}
\end{minipage}%
\hspace{1em}
\begin{minipage}{.46\columnwidth}
  \centering
  \includegraphics[scale=0.26]{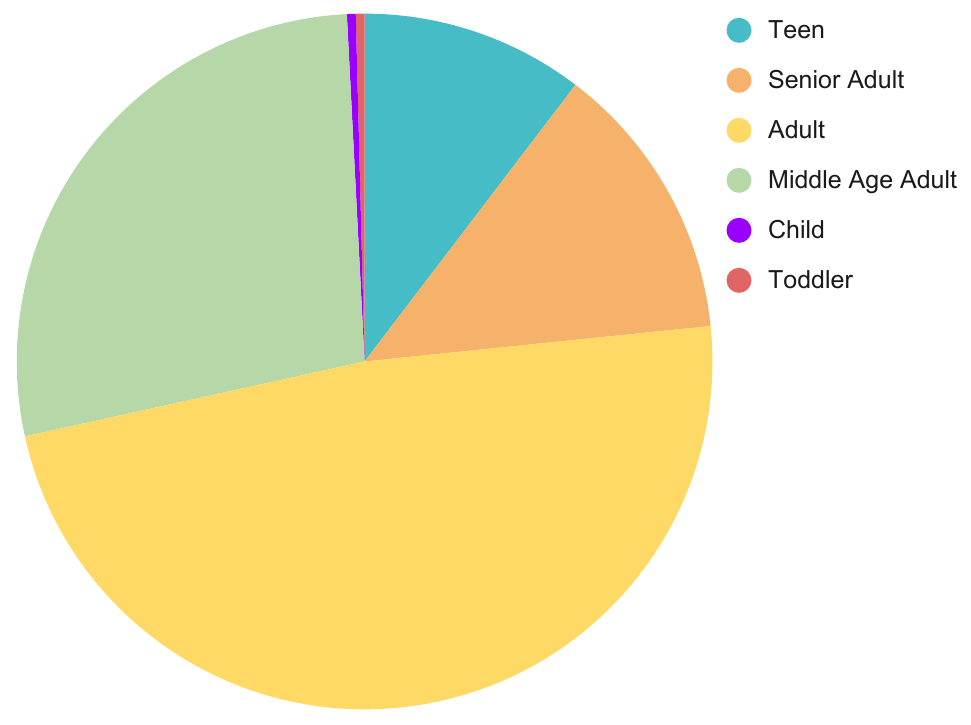}
  \captionof{figure}{Age group distribution}
  \label{AGE}
\end{minipage}
\vspace{-1.5em}
\end{figure} 

\hspace{-0.55cm}\textbf{Role of Patient's Personal Information} In real life, doctors often consider patients' personal context for determining a medical condition and plan treatment accordingly. Thus, we also annotated patients' personal information for each counseling session. Figures \ref{GEN} and \ref{AGE} demonstrate the distribution of gender and age among the patients in the dataset.

\section{Methodology}
\begin{figure*}[t]
    \centering
    \vspace{-1em}
    \includegraphics[width=\linewidth]{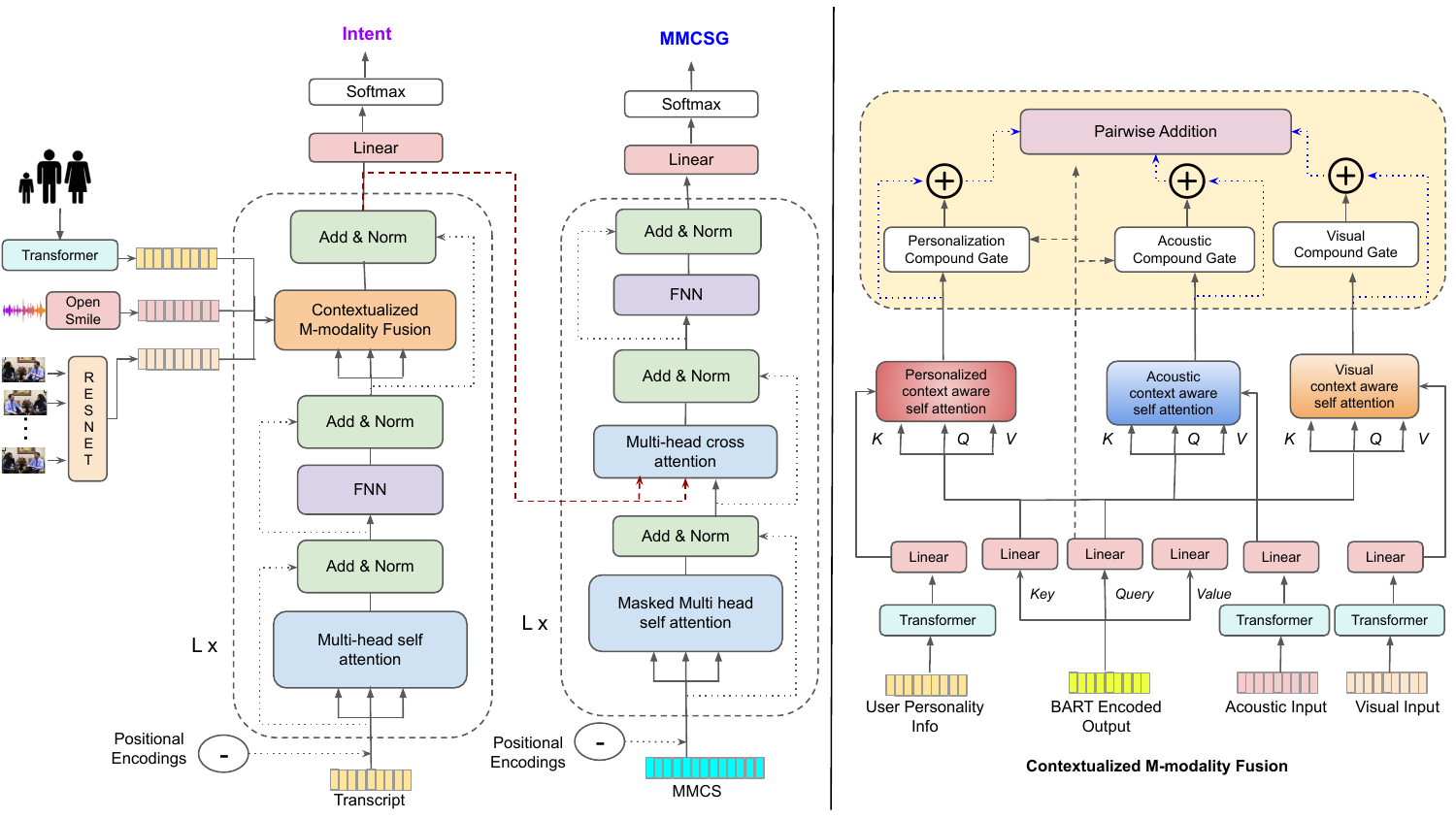}
    \vspace{-2.2em}
    \caption{Architecture of the proposed Multitasking multi-modal intent recognition and medical concern summary generation ({\em IR-MMCSG}) framework.}
    \vspace{-1.2em}
    \label{Model}
\end{figure*}

We anticipate that {\em MMCSG} is affected by (a) patient's visual expression, (b) patient's intent of communication, and (c) patient's personal information. Thus, we propose a multi-tasking, multi-modal intent recognition, and multi-modal medical concern summary generation ({\em IR-MMCSG}) framework. The proposed architecture is illustrated in Fig. \ref{Model}. There are three key stages: Multi-modal feature extraction, Contextualized M-modality fusion, and  Medical concern summary generation. The explanation and illustration of each stage and the flow are described below.
\subsection{Multi-modal Feature Extraction} In order to encode a counseling session, we have considered all three modalities of video, i.e., audio, text (transcript), and image (video frame). We extract the features of different modalities as follows:

\hspace{-0.55cm}\textbf{Textual Features:} The textual features of a therapy session represent the text embedding of its transcript. We use existing T5 \cite{raffel2020exploring} and BART \cite{lewis2019bart} tokenizers for transcript embedding. 

\hspace{-0.55cm}\textbf{Audio Features:} We extracted the audio feature from one of the most popular audio processing platforms named openSMILE \cite{eyben2010opensmile}. The feature representation considers maxima dispersion quotients, glottal source parameters, low-level descriptors (LLD), voice quality, MFCC, pitch, and their statistics. 

\hspace{-0.55cm}\textbf{Video Features:} We extracted frames from each counseling session at ten frames per second (fps). The frames extracted from the video are analyzed using Katna's approach \cite{wang2021toward}, aiming to identify frames with distinctive features. This process yielded a set of ten highly pertinent frames. Subsequently, these frames were passed to the pre-trained ResNet 152 model \cite{he2016deep} to acquire embeddings of these frames. Finally, we computed the average of these embeddings to form a representation vector of the video.

\subsection{Contextualized M-modality Fusion} We propose a novel multi-modal adapter-based infusion mechanism called contextualized M-modality fusion. It generates context- and modality-conditioned key and value vectors and produces a scaled dot product attention vector. The contextualized modality attention vector is utilized to calculate the global information attended over audio, visual, and personal context, which is utilized for intent identification and medical concern summary generation. It takes the hidden state (H) and calculates the contextualized modality attention as follows:
\begin{equation}
\small
    [Q K V] = H [W_Q, W_K, W_V]
\end{equation}

where $Q, K$, $V$ $\in \mathbb{R}^{l x d}$ are query, key, and value, respectively. Here, $l$ and $d$ denote the sequence length and the dimension of the hidden state (H), respectively. The term $W_Q, W_K$, and $W_v$ are the learnable parameters corresponding to the query, key, and value vectors, with the dimension of $\mathbb{R}^{d x d}$. 

To understand the medical consultation effectively, we generate different modalities and patients' personal context-conditioned key ($\hat{K}$) and value ($\hat{V}$) vectors. The attention vectors transpose the query vector (video transcript) to generate a contextualized, multi-modal, coherent information vector. The key and value pairs are calculated as follows: 
\begin{equation}
\small    {
\begin{bmatrix}
\hat{K} \\
\hat{V}
\end{bmatrix}} = {(1 - 
\begin{bmatrix}
{\lambda_k} \\
{\lambda_v}
\end{bmatrix}) {
\begin{bmatrix}
K \\
V
\end{bmatrix}} +
\begin{bmatrix}
{\lambda_k} \\
{\lambda_v}
\end{bmatrix}(M {
\begin{bmatrix}
{U_k} \\
{U_v}
\end{bmatrix}})}
\end{equation}
where $\lambda \in \mathbb{R}^{l x 1}$ is the learnable parameter that determines how much information from the textual modality should be retained and how much other modality information should be integrated. Here, $M$ denotes modality, which could be audio, video, and personal information. $U_k$ and $U_v$ are the learnable parameters. The modality controlling parameters ($\lambda$) are calculated using the gating mechanism as follows:
\begin{equation}
\small
     {
\begin{bmatrix}
\lambda_k \\
\lambda_v
\end{bmatrix}} = { \sigma ({
\begin{bmatrix}
K \\
V
\end{bmatrix}} {
\begin{bmatrix}
W_{k_1} \\
W_{v_1}
\end{bmatrix}} + M {
\begin{bmatrix}
{U_k} \\
{U_v}
\end{bmatrix}} {
\begin{bmatrix}
W_{k_2} \\
W_{v_2}
\end{bmatrix}}) }
\end{equation}
where $W_{k_1}, W_{k_2}, W_{v_1}$ and $W_{v_2}$ ($\in \mathbb{R}^{d x 1}$) are trainable weight matrices. Finally, the modality aware attentions ($H_a, H_v$ and $H_p$) and the final attended vector ($\hat{H}$) are calculated as follows:
\begin{equation}
\small
\centering
\begin{split}
& H_a  = Softmax (\frac{Q\hat{K}_{a}^{T}}{\sqrt{d_{k}}})\hat{V}_{a} \\
& H_v  = Softmax (\frac{Q\hat{K}_{v}^{T}}{\sqrt{d_{k}}})\hat{V}_{v} \\
& H_p = Softmax (\frac{Q\hat{K}_{p}^{T}}{\sqrt{d_{k}}})\hat{V}_{p}
\end{split}
\end{equation}

\hspace{-0.43cm}\textbf{Fusion} In order to infuse and control the amount of information transmitted from the different modalities (user personality information, audio, and visual), we build three compound gates: personalization ($g_p$), acoustic ($g_a$), and visual ($g_v$). The context information is transmitted via the gates as follows:
\vspace{-0.6em}
\begin{equation}
\centering
\begin{split}
& g_a = [H \oplus H_a]W_a + b_a  \\
& g_v  = [H \oplus H_v]W_v + b_v \\
& g_p = [H \oplus H_p]W_p + b_p 
\end{split}
\end{equation}

where $\oplus$ denotes a concatenation operation. $W_a, W_v, W_p$ ($\in \mathbb{R}^{2d X d}$) and $b_a$, $b_v$, $b_p$ ($\in R^{d X 1} $) are parameters. The final contextualized attended vector ($\hat{H}$) is computed as follows: 
\begin{equation}
    \hat{H} = H + g_p \odot	H_p + g_a \odot	H_a + g_v \odot	H_v 
    \label{ff}
 \end{equation}

\subsection{MMCS Generation}
MMCS is our primary task, which is being comprehended with the other task, intent recognition. We take the attended multi-modal encoder representation vector ($\hat{H}$) and pass it to a linear layer for intent identification. The vector ($\hat{H}$) is fed to the decoder's multi-head attention layer as key and value, with the key as the hidden representation of the medical concern summary. The infused information is processed with the traditional transformer's layers and computes the vocabulary's probability distribution. We have utilized a joint categorical cross-entropy loss function, which is the sum of loss functions of classification (CL) and generation (GL) tasks, i.e., $L = \alpha_1 * CL + \alpha_2 * GL  $ and $ \alpha_1 (=0.2) +  \alpha_2 (=0.8) =1$. 

\subsection{Experimental Details} We have utilized the PyTorch framework for implementing the proposed model. The generation models have been trained, validated, and evaluated with 80\%, 6\%, and 14\% samples of the {\em MM MediConSummation} dataset, respectively. The hyperparameter values, which are selected empirically, are as follows: sequence length (480), output max len (50), learning rate (3e-05), batch size (16), and activation function (ReLU). The different baselines and state-of-the-art models are listed as follows: 
\begin{itemize}
    \item \textbf{Seq2Seq-Transformer} It is a transformer-based sequence-to-sequence model \cite{shi2021neural}, which takes a combined representation of transcript, audio, and video features as input and generates a medical concern summary.
    \item \textbf{BART} It is a denoising autoencoder model that is trained to reconstruct corrupted sentences \cite{lewis2019bart}.
    \item \textbf{T5} It \cite{raffel2020exploring} is a versatile text-to-text model that combines encoder-decoder architecture with pre-training on a mixture of unsupervised and supervised tasks. 
    \item \textbf{MAF} MAF \cite{kumar2022did} is a fusion model that incorporates an additional adapter-based layer in the encoder of BART to infuse information from different modalities. 
    \item \textbf{MMCSG} is the proposed model with the proposed contextualized M-modality fusion mechanism only (without the multi-task intent recognition and MMCG generation setting).
    \item \textbf{{\em IR-MMCSG}} is the proposed multi-tasking, multi-modal intent identification, and medical concern summary generation model, incorporated with contextualized M-modality fusion mechanism. 
    \vspace{-1em}
\end{itemize}

\section{Result and Discussion}
 The purpose of the proposed multi-task framework is to enhance the performance of the primary task, MCSG, by utilizing the additional task of intent recognition. Thus, the results and analysis of MCSG are emphasized as the main focus in all task combinations.
\subsection{Experimental Results} The obtained performance by different multi-modal medical concern summary generation models are reported in Table \ref{R1}. Furthermore, we also investigated these models' efficacies for doctor summary generation, and results are presented in Table \ref{R2}. We ran experiments for ten iterations with different random seeds and reported the average values. The reported values in the following tables are statistically significant as the p-values obtained from Welch’s t-test \cite{welch1947generalization} at 5\% significance level are less than 0.05. 

\begin{table*}[hbt!]
    \caption{Performances of different models for multi-modal medical concern summary generation. Here, $^\dag$ indicates statistical significant findings ($p$ $<$ 0.05 at 5\% significance level).}
    \scalebox{0.75}{
    \begin{tabular}{|l|c|c|c|c|c|c|c|c|c|c|c|}
    \hline
     \textbf{Model}    & \textbf{BLEU-1} & \textbf{BLEU-2} & \textbf{BELU-3} &\textbf{BELU-4} & \textbf{BLEU}  & \textbf{ROUGE - 1}  & \textbf{ROUGE - 2} & \textbf{ROUGE- L}  & \textbf{METEOR} \\ \hline
     Seq2Seq-Transformer \cite{shi2021neural} &  18.65 & 10.93 & 3.85 & 1.23 & 8.67 & 24.03 & 8.62 & 22.16 & 21.36  \\ 
     T5 \cite{raffel2020exploring} & 19.67 & 11.73 & 5.92 & 1.93 & 9.81 & 28.23 & 11.01 & 26.48 & 23.44  \\
     BART \cite{lewis2019bart}  & 19.46 & 12.47 & 6.90 & 2.62 & 10.36 & 26.85 & 12.92 & 26.01 & 32.46 \\
     MAF \cite{kumar2022did} & 21.89 & 13.05 & 6.05 & 3.26 & 11.06 & 30.04 & 12.43 & 26.93 & 30.10 \\
     MMCSG w/o visual & 21.63  & 13.57 & 7.33 & 2.97 & 11.37 & 30.60 & 14.27  & 27.56 & 34.44 \\
     MMCSG & 23.26 & 14.13 & 7.80 & \textbf{3.81$^\dag$} & 12.25 & 31.47 & 14.16 & 28.51 & 33.60 \\
    \textbf{\em IR-MMCSG}  & \textbf{23.72$^\dag$}  & \textbf{14.68$^\dag$} & \textbf{7.88$^\dag$} & 2.98 & \textbf{12.31$^\dag$} & \textbf{32.16$^\dag$} & \textbf{14.41$^\dag$} & \textbf{29.61$^\dag$} & \textbf{35.87$^\dag$} \\ 
     \hline
    \end{tabular}}
    \label{R1}
    \vspace{-1em}
\end{table*}

\begin{table*}[hbt!]
    \centering
    \caption{Performances of different models for doctors' impression summary generation.  Here, $^\dag$ indicates statistical significant findings ($p$ $<$ 0.05 at 5\% significance level).}
    \scalebox{0.8}{
    \begin{tabular}{|l|c|c|c|c|c|c|c|c|c|c|c|}
    \hline
     \textbf{Model}    & \textbf{BLEU-1} & \textbf{BLEU-2} & \textbf{BELU-3} &\textbf{BELU-4} & \textbf{BLEU}  &  \textbf{ROUGE - 1}  & \textbf{ROUGE - 2}  & \textbf{ROUGE- L} & \textbf{METEOR}  \\ \hline
     T5 \cite{raffel2020exploring} & 17.37 & 9.18 & 2.51 & 0.088 & 7.28 & 26.70 & 7.83 & 23.60 & 18.44 \\
     BART \cite{lewis2019bart} & 20.87 & 9.96 & 1.22 & 0.417 & 8.12  & 26.86 & 8.04 & 22.84 & 23.35 \\
    MAF \cite{kumar2022did} & 20.88 & 11.53 & 3.69 & 2.69 & 9.70 & 26.99 & 9.83 & 24.03  & 23.89\\
    MMCSG w/o visual &  21.10 & 11.66 & 2.52 & 0.934  & 9.05 & 28.59 & 10.17 & 25.17 & 26.88 \\
    MMCSG &  23.28 & 12.38 & 3.71 & \textbf{3.13$^\dag$}  & 10.63 & 30.46 & 10.41 & 26.35 & 27.52 \\
    \textbf{\em IR-MMCSG} & \textbf{23.78$^\dag$} & \textbf{12.67$^\dag$} & \textbf{4.01$^\dag$} & 2.57 & \textbf{10.76$^\dag$} & \textbf{31.23$^\dag$} & \textbf{10.86$^\dag$} & \textbf{26.43$^\dag$} & \textbf{29.17$^\dag$}  \\
     \hline
    \end{tabular}}
    \label{R2}
     \vspace{-1em}
\end{table*}

\hspace{-0.55cm}\textbf{Ablation Study}  In order to understand the effectiveness of various components within the proposed model, we carried out an ablation study involving different combinations of these components. The obtained result has been reported in Table \ref{AS}. The results show that the model's performance is improving as the various different modalities are infused together to represent the context. 
\vspace{-2em}

\begin{table}[hbt!]
    \centering
    \caption{Performance of the proposed model with different modalities.  Here, $T$, $P$, $A$, and $V$ indicate transcript, personality information, audio, and video features, respectively. Here, $^\dag$ indicates statistical significant findings ($p$ $<$ 0.05 at 5\% significance level).}
    \scalebox{0.85}{
    \begin{tabular}{|l|c|c|c|c|c|}
    \hline
     \textbf{Model}  &  \textbf{BLEU} & \textbf{ROUGE-1} & \textbf{ROUGE-2} & \textbf{ROUGE- L}  & \textbf{METEOR} \\ \hline
     T & 11.00 & 28.88 &14.05 & 26.88 & 33.71  \\ 
    T + A & 11.37 & 30.60 & 14.27 & 27.56 & 35.71 \\
   T + P & 11.95 & 30.78 & 14.64 & 27.80 & 34.91 \\
     T + V & 11.73 & 30.89 & 14.44 & 28.18 & 35.80 \\
     T + A + V  & 11.56 & 30.46 & \textbf{14.84}$^\dag$ & 28.48  & 35.82  \\
    \textbf{T + A + V + P}  &  \textbf{12.31}$^\dag$ & \textbf{32.16}$^\dag$  & 14.41 & \textbf{29.61}$^\dag$  & \textbf{35.87}$^\dag$  \\
     \hline
    \end{tabular}}
    \label{AS}
    \vspace{-2em}
\end{table}

\hspace{-0.55cm}\textbf{Human Evaluation} We have also conducted a human evaluation of all test samples. In this assessment, two medical domain experts and one researcher (other than the authors) were employed to evaluate the generated medical concern summaries (twenty samples for each model) without revealing the models' names. The samples are assessed based on the following five metrics: \textit{domain relevance (DR), adequacy, fluency, informativeness (Info), and patient's personal context coherence (PC)} on a scale of 0 to 5. The obtained scores are presented in Table \ref{HE_Des}.

\begin{table}[hbt!]
    \centering
        \caption{Human evaluation for medical concern summary generation models}
    \scalebox{0.85}{
    \begin{tabular}{|l|c|c|c|c|c|c|}
    \hline
     \textbf{Model}  & \textbf{DR} & \textbf{Adequecy} & \textbf{Fluency} &  \textbf{PC} & \textbf{Info}  & \textbf{Avg.} \\ \hline
     Seq2Seq  \cite{shi2021neural} & 2.86 & 2.72 & 3.88 & 2.40 & 3.61 & 3.09\\
     BART \cite{lewis2019bart}    & 3.10 & 3.16 & 4.22 & 2.65 & 3.80 & 3.37 \\
     MAF \cite{kumar2022did} & 3.28 & 3.56 & 4.38 & 2.80 & 3.94 & 3.59 \\
     MMCSG & 3.44  & 3.81 & 4.55 & \textbf{3.14} & 3.84 & 3.76 \\
    \textbf{{\em IR-MMCSG}} & \textbf{3.82} & \textbf{3.94} & \textbf{4.74} & 3.08 & \textbf{4.06} & \textbf{3.93}\\
     \hline
    \end{tabular}}
    \vspace{-2em}
    \label{HE_Des}
\end{table}

\hspace{-0.55cm}\textbf{Key Observations} The main observations and insights are as follows: \textbf{(i)} The proposed medical concern summary generation model outperforms traditional sequence-to-sequence and transfer-learning-based generation models by a large margin, highlighting the importance of (a) task-specific conditioning and (b) the incorporated contextualized M-modality fusion (BART with early fusion- a simple concatenation of modalities vectors). \textbf{(ii)} For MCSG, visual modality (movements and expressions) infusion with text was more important than audio and demographic information (Table \ref{AS}). \textbf{(iii)} In human evaluation, we observed that the multi-task model incorporating intent representation generates a significantly more contextualized and informative summary of medical concerns (Table \ref{HE_Des} and Fig. \ref{CS}). 
\subsection{Findings to Research Questions}
\textbf{RQ 1:} {\em Do visuals and patient expression help in identifying patient key concern and generating adequate medical advice for the same?} The proposed MMCSG model outperforms the medical concern summary generation without visual information across all evaluation metrics (Table \ref{R1} and Table \ref{AS} - $T$ vs. $T$ + $V$ \& $T$ + $A$ vs. $T$ + $A$ + $V$). Furthermore, a similar trend has been observed for doctor suggestion summary generation (Table \ref{R2}). The enhanced findings strongly establish the effectiveness of utilizing visual cues, such as patients' facial expressions and body movements, in accurately identifying their medical conditions and responding accordingly. \\
\hspace{-0.36cm}\textbf{RQ 2:}{\em Can the patient's demographic context aid in generating an appropriate and relevant medical concern summary /suggestion?} Some behaviors, namely facial expressions/body movement and medical conditions, are heavily influenced by demographic information such as age and gender. Therefore, the proposed contextualized M modality fusion aided {\em IR-MMCSG} (Table \ref{R1} - MAF vs. MMCSG, Table \ref{R2} - MAF vs. MMCSG and Table \ref{AS} - $T$ vs. $T$ + $P$ \& $T$ + $A$ + $V$ vs. $T$ + $A$ + $V$ + $P$) that exploits the information to analyze and constrain different modalities performed significantly better than the non-context aware models.  \\
\hspace{-0.36cm}\textbf{RQ 3:} {\em Does any correlation exist between medical concern summary generation and user intent identification?} If a psychiatrist is aware of the reason for a patient's visit, it becomes more straightforward to identify the patient's primary mental issue and suggest him/her accordingly. The observed results firmly support the hypothesis, revealing that the multi-tasking {\em IR-MMCSG} clearly outperforms the single-task model, MMCSG (Table \ref{R1} and Table \ref{R2} - MMCSG vs. {\em IR-MMCSG}). Furthermore, we also observed a significant improvement by {\em IR-MMCSG} model in human evaluation (Table \ref{HE_Des}) across different metrics.
\section{Analysis} 
\begin{figure}[hbt!]
    \centering
    \includegraphics[scale=0.5]{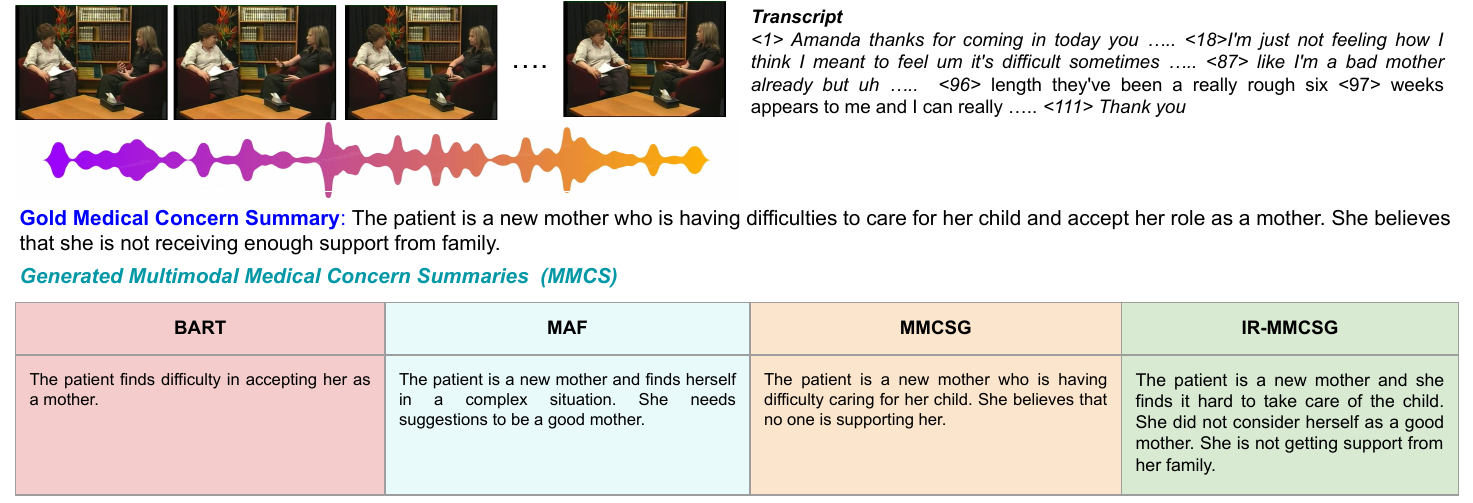}
    \vspace{-1.5em}
    \caption{Performances of different baselines and the proposed {\em IR-MMCSG} model for a test case}
    \vspace{-1em}
    \label{CS}
    \vspace{-3.7em}
\end{figure}

We have analyzed the proposed model's generated medical concern summaries and different models' behavior on some test cases, which are presented in Fig. \ref{CS}. The comprehensive analyses lead to the following key observations: \textbf(i) The proposed {\em IR-MMCSG} generates medical concern summaries (Fig. \ref{CS}) that include (a) a comprehensive and contextualized understanding of the patient's concerns and (b) a sense of the discussion that will be undertaken during the session. \textbf{(ii)} During the human evaluation, we found a significant number of cases where our model has also generated a cause of abnormality in the MCS (Table \ref{A1}). \textbf{(iii)} We observed that the models added additional words in the medical concern summary, leading to low evaluation scores despite being relevant and informative (Table \ref{A1}). 
\vspace{-1.5em}

 \begin{table}[hbt!]
    \centering
        \caption{Some generated medical concern summaries by the proposed IR-MMCSG model and their corresponding gold summaries.}
    \scalebox{0.84}{
    \begin{tabular}{|l|p{6.7cm}|p{6.7cm}|}
    \hline
   \textbf{No.} &  \textbf{Gold MMCS}   & \textbf{Generated MMCS}  \\ \hline
1 & The patient is feeling uncomfortable in his job.   & The patient is feeling increasingly uncomfortable in his {new job}. \\ 
  2 &  The patient has self-doubt about herself and she is worried about being judged.   & The patient feels depressed, she thinks that everyone is judging her, and it makes her upset. \\ 
  3 &  The patient realizes that he is overgeneralizing the fact that he is not going to fit in with their colleague. Meanwhile, he does not know much about his colleague. & The patient has the issue of feeling non-existing. He feels that he is different from other people at work. \\      \hline
    \end{tabular}}
    \label{A1}
        \vspace{-1.2em}
\end{table}

\hspace{-0.5cm}\textbf{Limitations} We have observed the following key limitations of the proposed medical concern summary generation model: \textbf{(i)} In the dataset, there were few instances where sets of symptoms and conditions were displayed on the screen, along with their status. Using OCR technology in these instances could lead to more accurate and precise summaries of the patient's medical condition. \textbf{(ii)} The transcripts of counselling sessions lack speaker type (psychiatrist and patient) information, which could have played an important role in effectively identifying and summarising patient concerns. \textbf{(iii)} Due to the limited demographic information (age and gender) and the small size of the dataset, there is a risk that the model may become biased towards certain age groups and gender that are more frequently represented.

\section{Conclusion} 
In this paper, we introduce a new task of generating a succinct summary of the primary concerns and expectations expressed by the initiator of a conversation. We curated the first multi-modal medical concern summary generation ({\em MM-MediConSummation}) corpus annotated with medical concern summary, user demographic information, user intent, and summary of doctors' suggestions. We proposed a multi-tasking multi-modal intent recognition and medical concern summary generation ({\em IR-MMCSG}) model incorporated with a novel adapter-based contextualized multi-modality fusion mechanism for analyzing acoustics and visual features with demographic and personality context. With the obtained results of various sets of experiments and human evaluation, we found firm evidence of the efficacy of the proposed {\em IR-MMCSG} model and the infused fusion mechanism over existing state-of-the-art methods. The improvements obtained establish the crucial role of facial expression/movement behavior and demographic context in identifying patients' medical concerns and generating an adequate summary for them. In the future, we would like to build an explainable multi-modal medical concern summary generation that generates medical concern summary generation along with evidence highlighting video spans. 

\section{Ethical Consideration}
 We strictly followed the medical research's ethical and regulatory guidelines particular to psychiatrist research \cite{avasthi2013ethics} during the dataset curation process. We have not added or removed any utterances/medical entity from the conversation. The curated dataset does not reveal users' identities, such as their names. The names have been replaced with some synthetic names. The annotation guidelines were provided by two psychiatrists, and the curated dataset is thoroughly checked and corrected by them. Furthermore, we have also obtained approval from our institute's healthcare committee and institutional ethical review board (ERB) to use the dataset and conduct the research. Thus, we confidently assert that the dataset, along with its comprehensive creation protocol, fully complies with the ethical and clinical imperatives of our discipline.

\section{Acknowledgment}
Abhisek Tiwari expresses sincere gratitude for being honored with the Prime Minister Research Fellowship (PMRF) Award by the Government of India. This grant has played a crucial role in supporting this research endeavor. Dr. Sriparna Saha extends sincere appreciation for receiving the Young Faculty Research Fellowship (YFRF) Award. This recognition is supported by the Visvesvaraya Ph.D. Scheme for Electronics and IT under the Ministry of Electronics and Information Technology (MeitY), Government of India, and executed by Digital India Corporation (formerly Media Lab Asia), has been invaluable in advancing the progress of the research. We want to express our gratitude to clinicians Dr. Pankaj Kumar (All India Institute of Medical Sciences Patna) and Dr. Minkashi Dhar (All India Institute of Medical Sciences Rishikesh) for their valuable contributions to this project.

%

\end{document}